\documentclass[10pt,twocolumn,letterpaper]{article}

\usepackage{cvpr}      %

\usepackage[dvipsnames]{xcolor}

\definecolor{cvprblue}{rgb}{0.21,0.49,0.74}
\usepackage[pagebackref,breaklinks,colorlinks,citecolor=cvprblue]{hyperref}
\usepackage{cuted}
\usepackage{multirow}
\usepackage{colortbl}
\usepackage{gensymb}

\def\paperID{3064} %
\def\confName{CVPR}
\def\confYear{2024}

\newcommand{\cjy}[1]{\textcolor{black}{#1}}

\definecolor{mygray}{rgb}{0.5,0.4,0.2}
\newcommand{\gray}[1]{\textcolor{mygray}{#1}}

\title{TextDiffuser-2: Unleashing the Power of Language Models for Text Rendering}

\author{%
  Jingye Chen$^{*13}$, Yupan Huang\thanks{Work done during internship at Microsoft Research.}\;$^{23}$, Tengchao Lv$^{3}$, Lei Cui$^{3}$, Qifeng Chen$^{1}$, Furu Wei$^{3}$ \\
  $^{1}$HKUST \;\;\;   $^{2}$Sun Yat-sen University \;\;\; $^{3}$Microsoft Research \\
  \texttt{qwerty.chen@connect.ust.hk, huangyp28@mail2.sysu.edu.cn, cqf@ust.hk} \\
  \texttt{\{tengchaolv,lecu,fuwei\}@microsoft.com}
}

\begin{document}

\maketitle

\begin{strip}
\centering
\vspace{-1.7cm}
\includegraphics[width=1\linewidth]{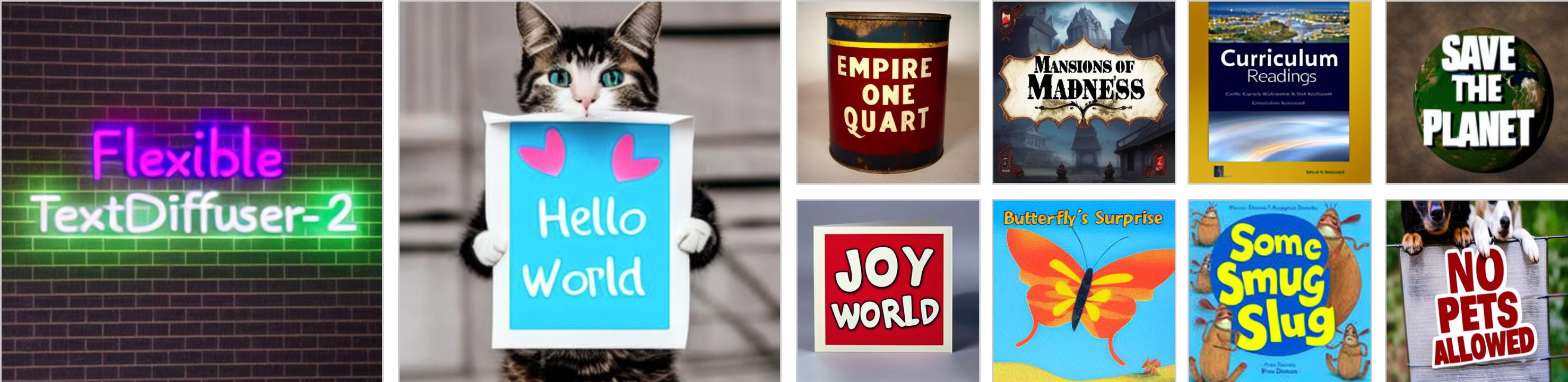}
\captionof{figure}{Text-to-image results generated by TextDiffuser-2. Alongside accurate text generation, TextDiffuser-2 offers reasonable text layouts and
exhibits diversity in text style powered by the strong capability of language models. \label{fig:teaser}}
\end{strip}


\begin{abstract}
    The diffusion model has been proven a powerful generative model in recent years, yet remains a challenge in generating visual text. Several methods alleviated this issue by incorporating explicit text position and content as guidance on where and what text to render. However, these methods still suffer from several drawbacks, such as limited flexibility and automation, constrained capability of layout prediction, and restricted style diversity. In this paper, we present TextDiffuser-2, aiming to unleash the power of language models for text rendering. Firstly, we fine-tune a large language model for layout planning. The large language model is capable of automatically generating keywords for text rendering and also supports layout modification through chatting. Secondly, we utilize the language model within the diffusion model to encode the position and texts at the line level. Unlike previous methods that employed tight character-level guidance, this approach generates more diverse text images. We conduct extensive experiments and incorporate user studies involving human participants as well as GPT-4V, validating TextDiffuser-2's capacity to achieve a more rational text layout and generation with enhanced diversity. The code and model will be available at \url{https://aka.ms/textdiffuser-2}.
\end{abstract}

\section{Introduction} \label{sec:intro}
In recent years, diffusion models \cite{ho2020denoising,rombach2022high,song2020denoising,gu2022vector,zhang2023adding,zhao2023uni,saharia2022palette} have successfully revolutionized the field of image synthesis, outperforming earlier methods based on GANs \cite{goodfellow2014generative,radford2015unsupervised} and VAEs \cite{kingma2013auto,rolfe2016discrete} in terms of fidelity and diversity. Despite showcasing impressive performance, most existing diffusion models still fall short in rendering visual text. Specifically, existing diffusion models often generate unintended symbols or artifacts during the text rendering process \cite{daras2022discovering}, which significantly impairs the visual quality of the generated images. Notably, text is ubiquitous in daily life, encompassing logos, banners, book covers, newspapers, etc. In this case, how to generate images with accurate, visually appealing, and coherent visual text is a crucial problem.

Through investigation, there has been a few research works focusing on visual text rendering \cite{chen2023textdiffuser,ma2023glyphdraw,yang2023glyphcontrol,liu2022character,balaji2022ediffi,saharia2022photorealistic,deepfloyd}. Some works \cite{saharia2022photorealistic,liu2022character,deepfloyd,balaji2022ediffi} validate that using powerful language models \cite{raffel2020exploring,xue2022byt5} as text encoders benefits the text rendering process. Nevertheless, they lack controllability since users may request to place text in a specific position. In this context, several works utilize explicit text position and content guidance, such as single-line segmentation masks and glyph images for GlyphDraw \cite{ma2023glyphdraw}, glyph images with multiple text lines for GlyphControl \cite{yang2023glyphcontrol}, as well as character-level segmentation masks for TextDiffuser \cite{chen2023textdiffuser}. Although showing impressive rendering accuracy, we have noticed several drawbacks in these methods: (1) \textbf{\textit{Limited flexibility and automation}}. GlyphControl \cite{yang2023glyphcontrol} needs users to design glyph images to provide layout guidance, while GlyphDraw \cite{ma2023glyphdraw} and TextDiffuser \cite{chen2023textdiffuser} rely on the manual specification of keywords. These requirements hinder the direct conversion of natural user prompts into corresponding images, thereby narrowing the flexibility and automation capabilities; (2) \textbf{\textit{Constrained capability of layout prediction}}. GlyphDraw \cite{ma2023glyphdraw} can only render images with a single text line, constraining its applicability for scenarios involving multiple text lines. For TextDiffuser \cite{chen2023textdiffuser}, the produced text layouts are not visually appealing, which is primarily attributed to the limited capability of the Layout Transformer; (3) \textbf{\textit{Restricted style diversity}}. For TextDiffuser \cite{chen2023textdiffuser}, the utilization of character-level segmentation masks as control signals implicitly imposes constraints on the position of each character, thereby restricting the diversity of text styles and posing challenges when rendering handwritten or artistic fonts.

Given these observations, we introduce TextDiffuser-2 in this paper, taking advantage of two language models for text rendering (samples are shown in Figure \ref{fig:teaser}). Firstly, we \textbf{\textit{tame a language model into a layout planner}} to transform user prompt into a layout using the caption-OCR pairs in the MARIO-10M dataset \cite{chen2023textdiffuser}. The language model demonstrates flexibility and automation by inferring keywords from user prompts or incorporating user-specified keywords to determine their positions. In addition, through chatting, users can guide the language model to alter the layout, such as regenerating, adding, or moving keywords. Secondly, we \textbf{\textit{leverage the language model in the diffusion model as the layout encoder}} to represent the position and content of text at the line level. Contrary to prior methods that utilized tight character-level guidance, this approach enables diffusion models to generate text images with broader diversity. Through comprehensive experiments and user studies that engaged both human participants and GPT-4V, we validate that TextDiffuser-2 can generate reasonable and visually pleasing text layouts, and it enhances the style diversity of the generated text. We will release the code and model to promote future research.

\section{Related work}

\paragraph{Visual text rendering.}
Despite the significant advancements in diffusion models \cite{ho2020denoising,rombach2022high,zhang2023adding,gu2022vector}, the generation of visual text rendering remains a persistent challenge. The advancement in visual text rendering will significantly enhance the efficiency of designers in executing text-related creative tasks, such as logo or poster design. Some work \cite{saharia2022photorealistic,liu2022character,balaji2022ediffi} leverage the large language models \cite{xue2022byt5,raffel2020exploring} to enhance the spelling capabilities of generative models. The other line of works \cite{chen2023textdiffuser,ma2023glyphdraw,yang2023glyphcontrol} attempts to explicitly control the position and content of the text to be rendered. For instance, GlyphDraw \cite{ma2023glyphdraw} comprises two diffusion models, including one for single-line text position prediction and the other one for image rendering guided by glyph images. GlyphControl \cite{yang2023glyphcontrol} utilizes glyph images with multiple text lines as prior to guide diffusion models render accurate and coherent text. Notably, TextDiffuser \cite{chen2023textdiffuser} uses character-level segmentation masks as more fine-grained signals for better rendering. Besides, TextDiffuser is a versatile model that can tackle text-to-image, text-to-image with template, and text inpainting tasks. TextDiffuser-2 falls in the latter line but distinguishes them by employing a language model for layout planning and using another language model to encode line-level text information to enable diverse rendering.

\paragraph{Language model for layout generation.}
Layout generation \cite{li2019layoutgan,jyothi2019layoutvae,gupta2021layouttransformer} has a wide range of applications, including document formatting \cite{patil2020read,he2023diffusion}, screen UI design \cite{deka2017rico}, and image synthesis \cite{feng2023layoutgpt,li2023gligen}. Previous methods \cite{jyothi2019layoutvae,li2019layoutgan} usually model layout generation as a regression task, representing bounding boxes using continuous coordinates. Recent advancements, such as Pix2Seq \cite{chen2021pix2seq,chen2022unified}, have explored alternative methods by treating coordinates as discrete language tokens. Another representative work, LayoutGPT \cite{feng2023layoutgpt}, carefully designs prompts to guide GPT-4 \cite{gpt4} generating formatted layout information to assist in image synthesis. Recently, some multimodal large language models \cite{peng2023kosmos,lv2023kosmos,zhou2023regionblip,zhang2023gpt4roi,chen2023position,you2023ferret} have also adopted this design for grounding specific objects in images. In line with these designs, TextDiffuser-2 aims to leverage language models as layout planners for visual text rendering. 

\paragraph{Optical character recognition.}
Images naturally serve as carriers of textual information. Optical Character Recognition (OCR) has been extensively studied in academia. Specifically, text recognition \cite{shi2016end,yu2021benchmarking,li2023trocr,bautista2022scene,yu2023chinese} and detection \cite{ma2018arbitrary,zhou2017east,lyu2018multi,he2021most,liao2020real} techniques play a crucial role, aiming to locate and extract textual information and further facilitate high-level understanding tasks. Our method leverages caption-OCR pairs \cite{chen2023textdiffuser} to fine-tune a large language model for generating visual text layouts. Additionally, we employ OCR tools to conduct a comprehensive evaluation.

\begin{figure*}[ht]
\centering
\includegraphics[width=1.0\textwidth]{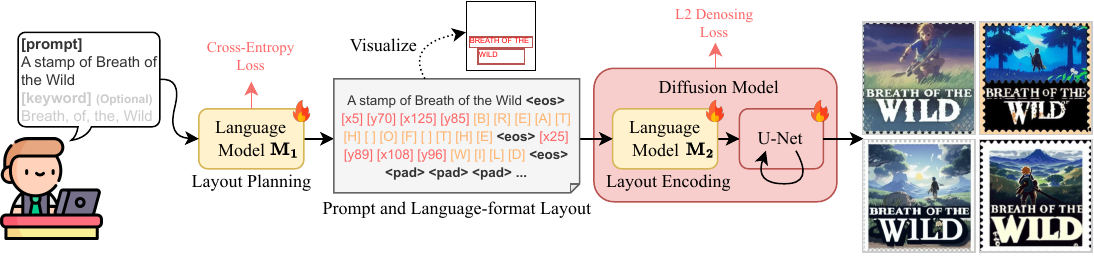}
\caption{
The architecture of TextDiffuser-2. The language model $\mathbf{M_{1}}$ and the diffusion model are trained in two stages. The language model $\mathbf{M_{1}}$ can convert the user prompt into a language-format layout and also allows users to specify keywords optionally. Further, the prompt and language-format layout is encoded with the trainable language model $\mathbf{M_{2}}$ within the diffusion model for generating images. $\mathbf{M_{1}}$ is trained via the cross-entropy loss in the first stage, while $\mathbf{M_{2}}$ and U-Net are trained using the denoising L2 loss in the second stage.}
\label{fig:model_arch}
\end{figure*}

\section{Methodology}

The architecture of TextDiffuser-2 is depicted in Figure \ref{fig:model_arch}, where the language model $\mathbf{M_{1}}$ and the diffusion model are trained in two stages. We introduce the role of two language models, including a language model for layout planning and another language model for layout encoding. We focus on introducing the text-to-image process, while the functions of text-to-image with templates and text inpainting will be introduced in the experiment section.

\subsection{Language model for layout planning}

Recent research has revealed that benefiting from the extensive training data across various domains, large language models \cite{gpt4,touvron2023llama,touvron2023llama2} exhibit expertise beyond the language domain, such as layout planning \cite{feng2023layoutgpt,lin2023layoutprompter}. Inspired by this, we try to tame a large language model into a layout planner.

Specifically, we seek to fine-tune a pre-trained large language model $\mathbf{M_{1}}$, which functions as a decoder, using caption-OCR pairs. As demonstrated in Figure \ref{fig:model_arch}, we consider two options: (1) If users do not explicitly provide keywords, the language model should infer the text and layout to be drawn on the image; (2) If users provide keywords (marked in gray color), the language model only needs to determine the corresponding layout for the keywords. Specifically, the input follows the format ``$\mathbf{[description]}$ Prompt: $\mathbf{[prompt]}$ {\color{gray}Keywords: $\mathbf{[keywords]}$}''\footnote{Task description: Given a prompt that will be used to generate an image, plan the layout of visual text for the image. The size of the image is 128x128. Therefore, none of the properties of the positions should exceed 128, including the coordinates of the top, left, right, and bottom. You don’t need to specify the details of font styles. At each line, the format should be textline left, top, right, and bottom. So let us begin.}. For the output, we expect each line to follow the format ``\textit{textline} $x_{0}$, $y_{0}$, $x_{1}$, $y_{1}$'', where ($x_{0}$, $y_{0}$) and ($x_{1}$, $y_{1}$) represent the coordinates of the top-left corner and bottom-right corner, respectively. We optimize the language model with cross-entropy loss, training simultaneously for scenarios with and without keywords. We use all the text detected in the OCR results as keywords to formulate the input. Moreover, we expect the fine-tuned language model can be guided to alter the generated layout through chatting. We will delve into this aspect in the discussion section.

\subsection{Language model for layout encoding}

Based on the layouts generated by $\mathbf{M_{1}}$, we leverage the latent diffusion models \cite{rombach2022high} for image generation. Different from TextDiffuser \cite{chen2023textdiffuser} which incorporates text information using segmentation masks and GlyphControl \cite{yang2023glyphcontrol} which duplicates backbone parameters to accommodate the glyph image conditions, we introduce a simple and parameter-free strategy by combining the prompt and the layout for the language model $\mathbf{M_{2}}$, \textit{i.e.}, the text encoder within the latent diffusion model. In contrast to character-level segmentation masks that regulate the position of individual characters, the line-level bounding box offers greater flexibility during generation and does not constrain the diversity of styles.

Previous work \cite{liu2022character} demonstrates that fine-grained tokenization can enhance the spelling capability of diffusion models. Inspired by this, we design a hybrid-granularity tokenization method that not only improves the spelling capability of the model but also keeps the sequence from getting too long. Specifically, on the one hand, we maintain the original BPE tokenization method \cite{sennrich2015neural} for the prompt. On the other hand, we introduce new character tokens and decompose each keyword into the character-level representation. For example, the word ``WILD'' is decomposed into tokens ``[W]'', ``[I]'', ``[L]'', ``[D]''. Additionally, we introduce new coordinate tokens to encode the position. For instance, the tokens ``[x5]'' and ``[y70]'' correspond to an x-coordinate of 5 and a y-coordinate of 70, respectively. Each keyword information is separated by the end-of-sentence token ``$\langle \text{eos} \rangle$'', and any remaining space to the maximum length $L$ will be filled with padding tokens ``$\langle \text{pad} \rangle$''. We train the whole diffusion model, including the language model $\mathbf{M_{2}}$ and U-Net, using the L2 denoising loss.

\section{Experiments}

\paragraph{Implementation details.}
For \textbf{\textit{layout planning}}, we fine-tune the vicuna-7b-v1.5 \cite{vicuna2023} model based on the FastChat framework \cite{zheng2023judging}. The caption-OCR pairs are derived from the MARIO-10M dataset \cite{chen2023textdiffuser}, and we use 5k samples for fine-tuning. We normalize the positions to the range of 0$\sim$128 to increase the compactness of the coordinate feature space. The learning rate is set to 2e-5, and we conduct a total of 6 epochs of fine-tuning with a batch size of 256. It takes one day to train with 8 A100 GPU cards. During the inference stage, when using a single A100 GPU card, the average time to generate a layout for each prompt is 1.1 seconds. For \textbf{\textit{layout encoding}}, we utilize SD 1.5 \cite{rombach2022high} and use the built-in CLIP text encoder with base size \cite{radford2021learning}. The whole model consists of 922M parameters. We incorporate special tokens, including 256 coordinate tokens and 95 character tokens. The alphabet contains 26 uppercase and 26 lowercase letters, 10 numbers, 32 punctuation marks, and a space. The size of the input image is 512$\times$512. The model is trained for 6 epochs on the MARIO-10M dataset \cite{chen2023textdiffuser} with a learning rate of 1e-4 and a batch size of 576. The maximum length $L$ is set to 128. More details about the choice of $L$ are shown in Appendix A. It takes one week to train the whole diffusion model with 8 A100 GPU cards.
When sampling with 50 steps, the generation for a single image costs 6 seconds.

\begin{table}[t]
\centering
\scalebox{1}{
\begin{tabular}{c c c c c c}
\toprule
\#Data & Acc${\uparrow}$ & Pre${\uparrow}$ & Rec${\uparrow}$ & F${\uparrow}$ & IOU${\downarrow}$ \\
\midrule
0k-2shot & 49.65 & 84.18 & 69.69 & 76.25 & 19.69 \\
\arrayrulecolor{gray}
\midrule
\arrayrulecolor{black}
2.5k & 61.10 & 82.20 & 85.18 & 83.67 & \textbf{3.21} \\
5k & \textbf{64.85} & 84.98 & \textbf{86.38} & \textbf{85.67} & 3.25 \\
10k & \textbf{64.85} & 84.38 & 86.23 & 85.29 & 4.27 \\
50k & 63.72 & \textbf{85.32} & 85.78 & 85.55 & 3.68 \\
100k & 62.87 & 85.26 & 85.98 & 85.62 & 4.31 \\

\bottomrule 
\end{tabular}}
\caption{Ablation studies on the amount of fine-tuning data. The ``0k-2shot'' setting denotes the use of two examples for few-shot learning, without any additional fine-tuning. When using 5k data, the language model $\mathbf{M_{1}}$ performs better. The percentage sign is omitted, as is consistent with the following tables. `Pre', `Rec', and `F' denote precision, recall, and f-measure, same as follows.}
\label{tab:data_amount}
\end{table}

\begin{table}[t]
\centering
\scalebox{1}{
\begin{tabular}{c c c c c}
\toprule
Representation & Acc${\uparrow}$ & Pre${\uparrow}$ & Rec${\uparrow}$ & F${\uparrow}$ \\
\midrule
Center (Char) & 35.19 & 61.75 & 62.71 & 62.23 \\
LT (Char) & 28.32 & 54.94 & 55.64 & 55.29 \\
\arrayrulecolor{gray}
\midrule
\arrayrulecolor{black}
LT+RB (Subword) & 15.48 & 41.74 & 42.53 & 42.13 \\
LT+RB (Char) & \textbf{57.58} & \textbf{74.02} & \textbf{76.14} & \textbf{75.06} \\
\bottomrule 
\end{tabular}}
\caption{Ablation studies on the representation of coordinates and the tokenization level. `L', `T', `R', and `B' denote left, top, right, and bottom. ``Char'' refers to tokenizing keywords into individual characters, whereas ``Subword'' refers to the use of BPE for tokenizing into subwords. Using the top-left and bottom-right corners and character-level tokenization achieves better performance.}
\vspace{-0.2cm}
\label{tab:representation}
\end{table}

\subsection{Ablation studies}

\begin{figure}[t]
\centering
\includegraphics[width=0.48\textwidth]{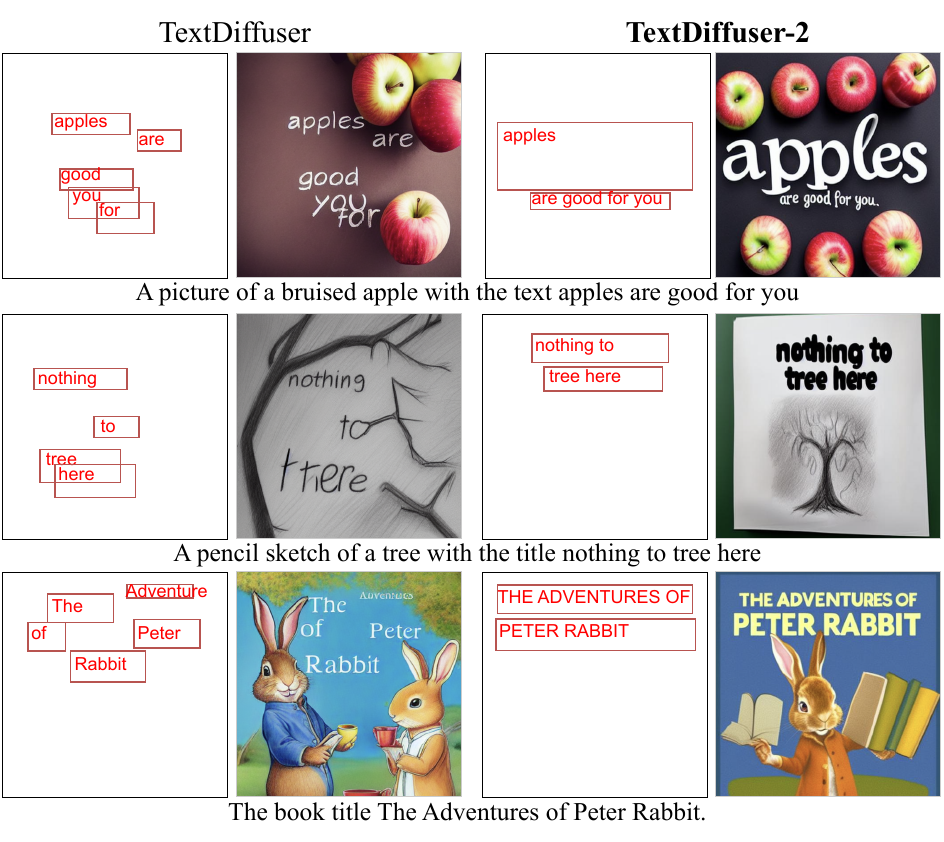}
\caption{
Visualizations of layouts. TextDiffuser-2 generates more visually pleasing and rational layouts compared with TextDiffuser.
}
\vspace{-0.2cm}
\label{fig:layout}
\end{figure}

\paragraph{How much data is needed for fine-tuning $\mathbf{M_{1}}$?} As illustrated in Table \ref{tab:data_amount}, we conduct experiments with different data amount, including 0k, 2.5k, 5k, 10k, 50k, and 100k. Particularly, in the 0k setting, we provide two examples of few-shot learning. In the absence of examples, the result often fails to conform to the appropriate format. We evaluate our approach using the MARIO-Eval benchmark \cite{chen2023textdiffuser}, which consists of prompt and keyword pairs. Besides, the quotation marks in the prompt are removed for evaluation. Since the LAIONEval subset contains some noise in its keyword ground truth, it is unreliable for accurate assessments. So, we decided not to use it in this keyword extraction experiment. For evaluation, we use accuracy, precision, recall, and F-measure to assess the model's ability to extract keywords. Additionally, we introduce an IoU metric to measure the maximum IoU value between the generated boxes for each sample (only those samples with more than one predicted box are calculated). The experimental results showcase that the model achieved optimal performance in the majority of metrics when fine-tuned with 5k data, and we visualize some samples in Figure \ref{fig:layout}. We notice that the language model exhibits flexibility in generating keywords, such as determining the case of the keyword, or introducing appropriate words beyond the provided prompt. More samples are in Appendix B. In subsequent experiments, we employ the model fine-tuned on 5k data for layout planning.

\begin{table*}[t]
\centering
\begin{tabular}{cccccc}
 \toprule
Metrics      & SD-XL \cite{podell2023sdxl}  & PixArt-$\alpha$ \cite{chen2023pixart} & GlyphControl \cite{yang2023glyphcontrol} & TextDiffuser \cite{chen2023textdiffuser} & TextDiffuser-2 \\
\midrule
\multicolumn{6}{l}{\textbf{\textit{Quantitative Results}}\;\;\;\;\;} \\
FID${\downarrow}$ & 62.54  & 87.09 & 50.82 & \underline{38.76} & \textbf{33.66} \\
CLIPScore${\uparrow}$  & 31.31 & 27.88 & \textbf{34.56} & 34.36 & \underline{34.50} \\
OCR (Accuracy)${\uparrow}$  & 0.31 & 0.02 & 32.56 & \underline{56.09} & \textbf{57.58} \\
OCR (F-measure)${\uparrow}$ & 3.66 & 0.03 & 64.07 & \textbf{78.24} & \underline{75.06} \\

\arrayrulecolor{gray}
\midrule
\arrayrulecolor{black}

\multicolumn{6}{l}{\textbf{\textit{User Studies by Humans / \gray{GPT-4V} }}\;\;\;\;\;} \\

Layout Aesthetics${\uparrow}$ & - & - & - & \underline{28.43} / \gray{\underline{0.00}} & \textbf{71.57} / \gray{\textbf{100.00}} \\

Style Diversity${\uparrow}$ & - & - & \underline{31.37} / \gray{\textbf{33.33}} & 27.45 / \gray{\textbf{33.33}} & \textbf{41.18} / \gray{\textbf{33.33}} \\

Text Quality${\uparrow}$ & 14.58 / \gray{7.69} & 3.65 / \gray{0.00} & 21.35 / \gray{15.38} & \underline{23.44} / \gray{\underline{30.77}} & \textbf{36.98} / \gray{\textbf{46.15}} \\

Text-Image Matching${\uparrow}$ & 7.14 / \gray{0.00} & 3.30 / \gray{0.00} & \underline{29.67} / \gray{18.18} & 19.23 / \gray{\underline{36.36}} & \textbf{40.66} / \gray{\textbf{45.45}} \\

Inpainting Ability${\uparrow}$ & - & - & - & \underline{25.49} / \gray{\underline{33.33}} & \textbf{74.51} / \gray{\textbf{66.67}} \\

\bottomrule
\end{tabular}
\caption{Demonstration of the quantitative results and user studies. We also incorporate GPT-4V \cite{gpt4} into the user studies. The best and second-best results are indicated in bold and underlined formats. TextDiffuser-2 achieves the best results under the majority of metrics.}
\label{tab:performance}
\end{table*}

\paragraph{How to represent the position of text lines?}
Apart from utilizing the top-left and bottom-right corners to represent a text line, we also investigate alternative single-point representations, such as employing the top-left point or the center point. Intuitively, using a single point to represent a text line provides more flexibility, enabling the generated text to exhibit greater diversity in angles and sizes. In \cjy{Appendix C}, visualizations are shown to validate the diversity of the generation using single-point conditions. However, as shown in Table \ref{tab:representation}, we notice that there is a considerable decline in the OCR accuracy of the single-point representation on the MARIO-Eval benchmark \cite{chen2023textdiffuser}. For example, compared with the LT-RB setting, the accuracy of the center and LR settings declined by 22.39\% and 29.26\%. Hence, we leverage the top-left and bottom-right corners to represent the box in the following experiments. We also explore the inclusion of angle information in \cjy{Appendix D}.

\paragraph{Should text be tokenized at the character or subword level?} We also explore Byte Pair Encoding (BPE) to tokenize keywords into the subword level.
As shown in Table \ref{tab:representation}, we observe that using subword-level tokenization significantly underperforms character-level representation, \textit{i.e.}, it is lower by 42.1\% on the accuracy metric. When using subword-level tokenization, the model becomes insensitive to the spelling of each token, which poses significant challenges to the text rendering process.

\subsection{Experimental results}

\paragraph{Quantitative results.} 

\begin{figure*}[t]
\centering
\includegraphics[width=1.0\textwidth]{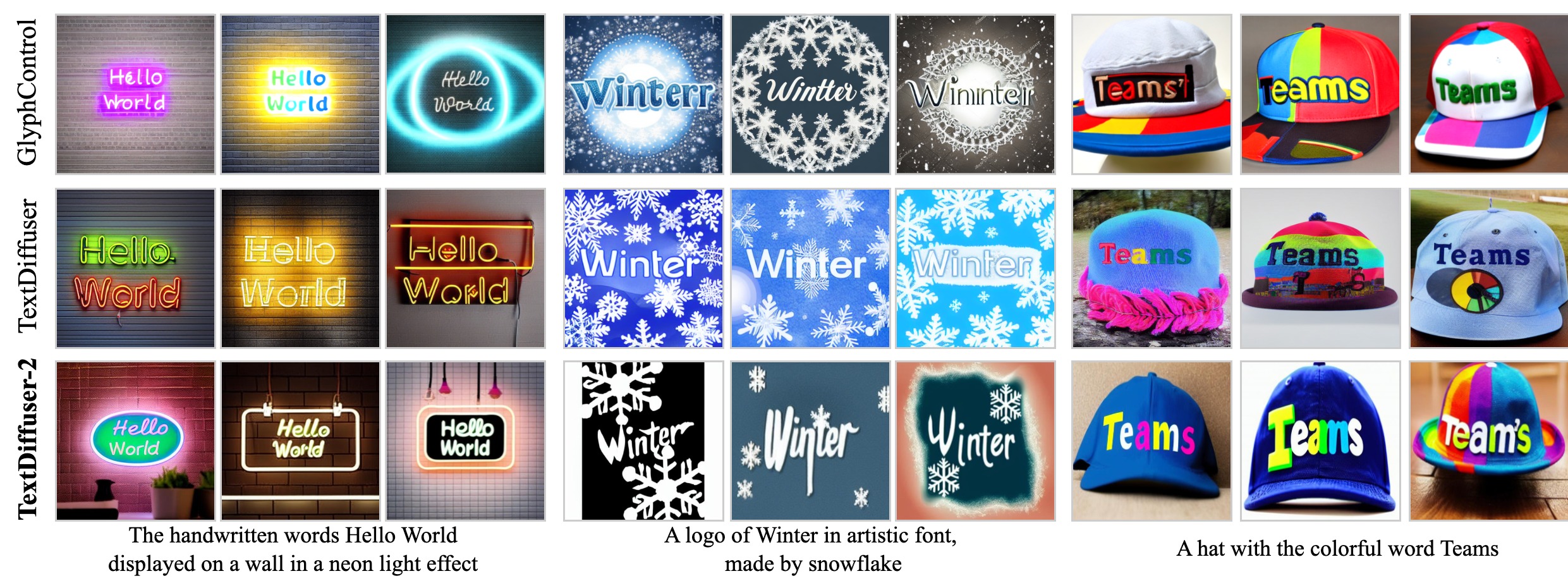}
\caption{
Visualization of diversity in generating multiple images under the same prompt. TextDiffuser-2 is capable of generating more artistic fonts, with increased diversity in the positioning of characters and the inclination angle of text lines.}
\label{fig:diversity}
\vspace{-0.2cm}
\end{figure*}

As shown in Table \ref{tab:performance}, we conduct quantitative experiments on the MARIO-Eval benchmark \cite{chen2023textdiffuser}. For comparisons, we leverage two state-of-the-art text-to-image models including SD-XL \cite{podell2023sdxl} and PixArt-$\alpha$ \cite{chen2023pixart}, and two models incorporating specific guidance for generating text images including TextDiffuser \cite{chen2023textdiffuser} and GlyphControl \cite{yang2023glyphcontrol}. Details of these compared methods are shown in \cjy{Appendix E}. For all methods, we employ 50 sampling steps and set the classifier-free guidance to 7.5. The experimental results demonstrate that TextDiffuser-2 outperforms other methods in terms of the FID evaluation metric. Having not been specifically trained on text images, SD-XL and PixArt-$\alpha$ exhibit a larger divergence against the ground truth, resulting in higher FID and lower CLIP scores. For the OCR metrics, it is observed that only models incorporating guidance can effectively render text. Furthermore, TextDiffuser-2 outperforms GlyphControl and has an OCR performance comparable to TextDiffuser. It is noteworthy that the TextDiffuser renders text in a standardized font (see Figure \ref{fig:diversity}), thereby reducing the complexity of the rendering process. This strategy sacrifices font style diversity to enhance the accuracy of text rendering. By contrast, while maintaining the ability to generate accurate text, TextDiffuser-2 can generate text with a greater diversity.

\begin{figure*}[ht]
\centering
\includegraphics[width=1.0\textwidth]{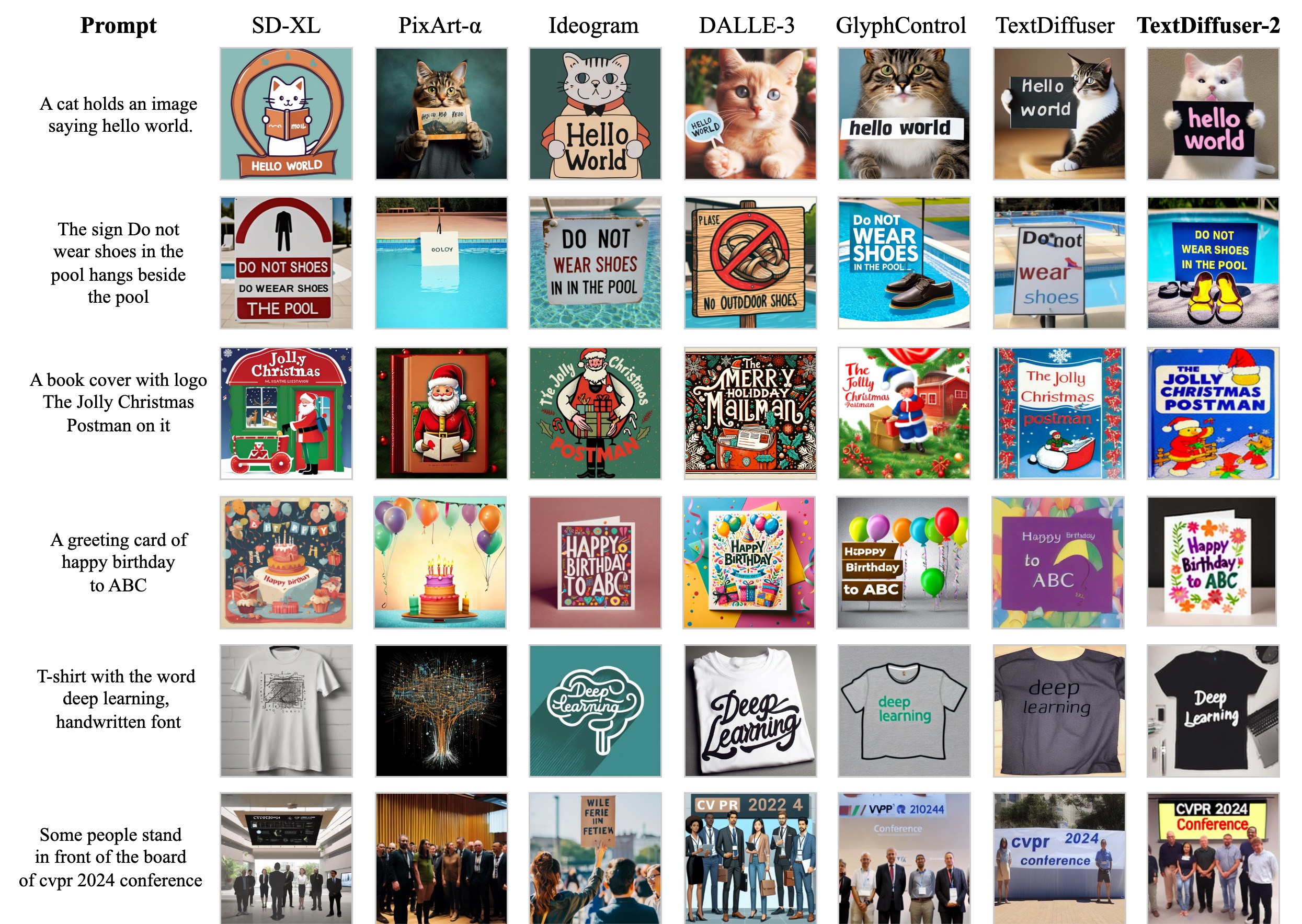}
\caption{
Visualizations of text-to-image results compared with existing methods. TextDiffuser-2 can automatically extract keywords from prompts for accurate rendering. Additionally, the fonts generated by TextDiffuser-2 exhibit a wide range of diversity.
}
\vspace{-0.2cm}
\label{fig:result}
\end{figure*}

\paragraph{Qualitative results.} 
The visualizations are demonstrated in Figure \ref{fig:result}. We compare our method with SD-XL \cite{podell2023sdxl}, PixArt-$\alpha$ \cite{chen2023pixart}, Ideogram \cite{ideogram}, DALLE-3 \cite{dalle3}, GlyphControl \cite{yang2023glyphcontrol} and TextDiffuser \cite{chen2023textdiffuser}. Although some of the latest text-to-image models (\textit{e.g.}, DALLE-3 and PixArt-$\alpha$) showcase superior image quality, they do not perform as well in rendering text compared with models that incorporate explicit guidance. Compared to TextDiffuser, our method generates more aesthetically pleasing layouts, avoiding misalignment or discordant font sizes. Furthermore, since TextDiffuser-2 utilizes the more flexible line-level guidance, it offers better control over font style, such as when rendering the handwritten style ``deep learning''. In contrast, TextDiffuser, which uses character-level guidance, can mainly render rigid font styles. We offer some visualizations in Figure \ref{fig:diversity}. For instance, when rendering ``Winter'', our method demonstrates greater diversity in terms of perspective angle and font style compared to other methods. In addition, we adopt the same layout to validate the performance of GlyphControl, which also uses line-level guidance. We observe that TextDiffuser-2 achieves a higher accuracy than GlyphControl. We also conduct comparisons with some methods that are neither open-source nor offer APIs, such as GlyphDraw \cite{ma2023glyphdraw} and Character-Aware Model \cite{liu2022character} using the samples shown in their corresponding papers in \cjy{Appendix F}.

 \paragraph{User studies.} As shown in Table \ref{tab:performance}, we design questions covering five aspects: layout aesthetics, style diversity, text quality, text-image matching, and inpainting ability, each of which contains 6, 3, 6, 6, 6 questions. We involved a total of 17 human participants in our study. Additionally, we employ GPT-4V \cite{gpt4} to carry out the user study. We devised instruction guidance for each task, prompting GPT-4V to think step by step to arrive at the final answer. Ultimately, each method's score is calculated as the number of votes it receives divided by the total number of votes. Based on the results, TextDiffuser-2 has achieved optimal performance in four out of five metrics in studies involving human participation and GPT-4V. More details are in \cjy{Appendix G}.

\begin{figure*}[t]
\centering
\includegraphics[width=1.0\textwidth]{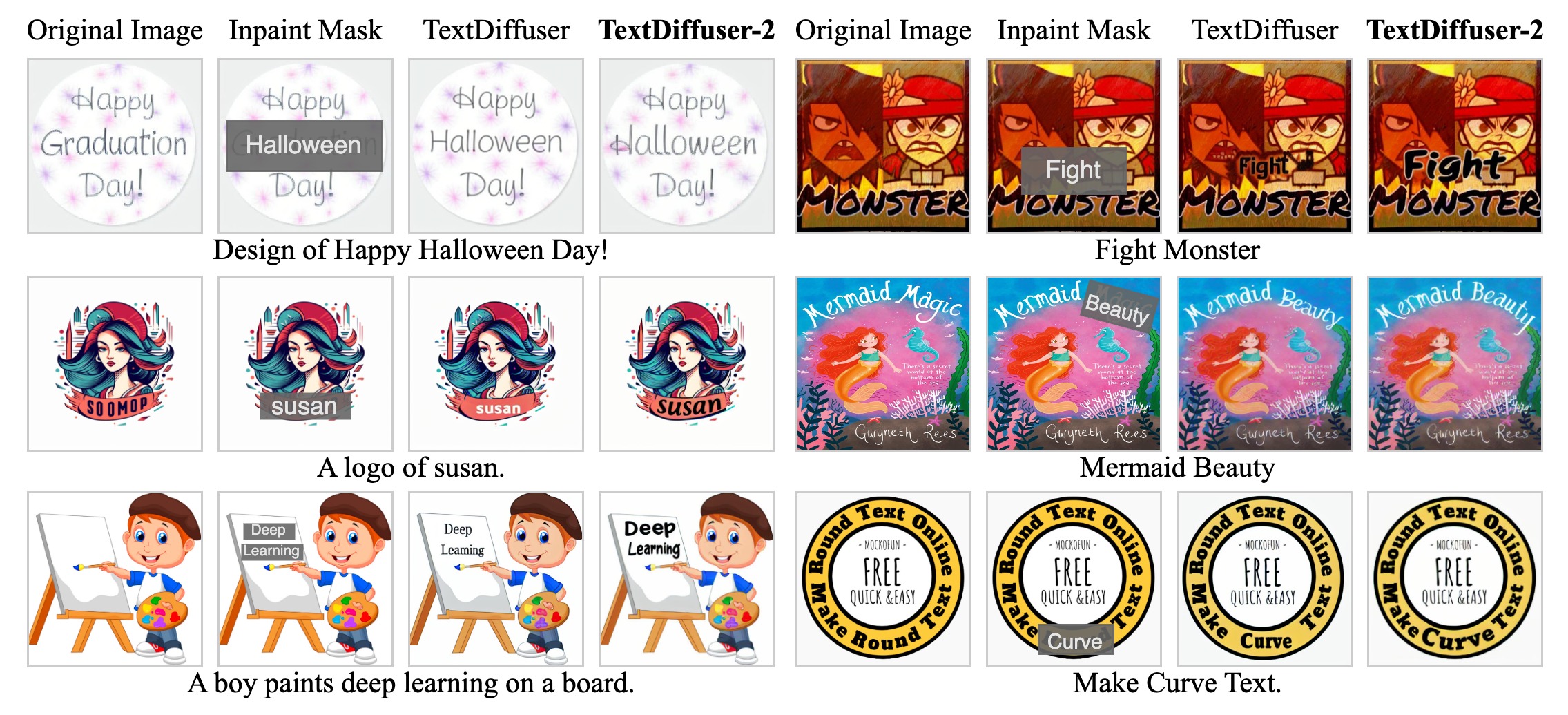}
\caption{
Visualizations of the text inpainting task compared with TextDiffuser. TextDiffuser-2 can generate more coherent text.
}
\label{fig:inpaint}
\vspace{-0.2cm}
\end{figure*}

\begin{figure}[t]
\centering
\includegraphics[width=0.476\textwidth]{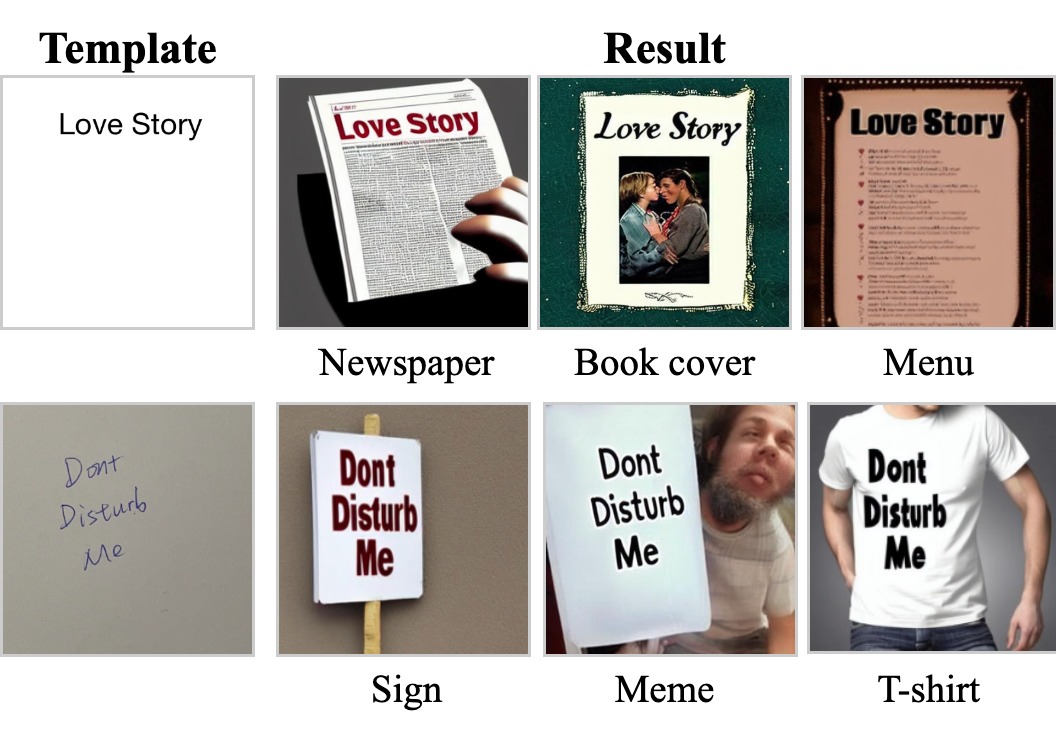}
\caption{
Visualizations of the text-to-image with template task.
}
\label{fig:template}
\end{figure}

\begin{figure}[t]
\centering
\includegraphics[width=0.49\textwidth]{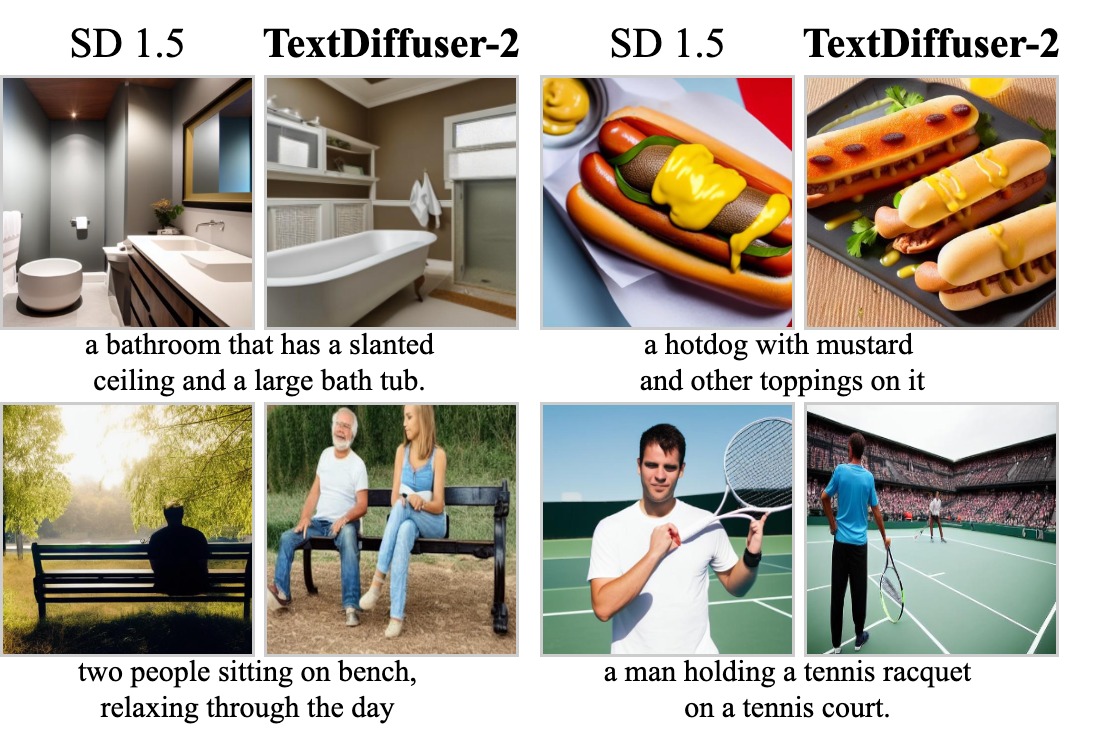}
\caption{
Visualizations of generating images without text.
}
\label{fig:general}
\end{figure}

\paragraph{More applications.} In addition to text-to-image, we also explore other applications of TextDiffuser-2. (1) \textit{\textbf{Text-to-image generation with template.}} When a template image (\textit{e.g.}, printed, handwritten, or scene text image) is provided, TextDiffuser-2 can use existing OCR tools to extract text information and directly feed it into the diffusion model as the condition, eliminating the need for layout prediction from the language model $\mathbf{M_{1}}$. We showcase some samples in Figure \ref{fig:template}. (2) \textit{\textbf{Text inpainting.}}
Similar to TextDiffuser, the architecture of TextDiffuser-2 adapts well for training on text inpainting tasks. We only need to modify the channel of the input convolution kernel in the U-Net. Specifically, we augment the original 4-dimensional latent feature with 5 additional dimensions, including 4 dimensions of non-inpaint area features and 1 dimension for the mask. Moreover, only the text position and content from the inpaint area are required as conditions for the diffusion model. More details are shown in \cjy{Appendix H}. We compare the performance with TextDiffuser, and the experimental results are shown in Figure \ref{fig:inpaint}. It is worth noting that TextDiffuser requires a text mask as a condition to specify the position of each character, which can be cumbersome in practical applications. Additionally, the text mask may limit the style of the generated results. For example, when rendering the word ``Curve'', the generated result cannot produce a visually curved effect due to the constraints of the character-level segmentation mask. In contrast, the inpainting process of TextDiffuser-2 is more flexible, thus resulting in a better user experience. (3) \textit{\textbf{Natural image generation without text.}} Given that we train TextDiffuser-2 on images with text, we are curious about its capability to generate images without text. Specifically, by omitting the text position and content guidance, TextDiffuser-2 can generate images without text. We randomly select 10,000 prompts from the Microsoft COCO dataset \cite{lin2014microsoft} for generation and compare the results with those generated by SD 1.5 \cite{rombach2022high}. The visualization results are shown in Figure \ref{fig:general}. Although TextDiffuser-2 is fine-tuned on domain-specific data, it still maintains its generative capabilities in the original domain. We calculate the FID score of the generated results. When the sampling steps are set to 50, and the classifier-free guidance is set to 7.5, TextDiffuser-2's FID score is 24.06, versus 23.03 for SD 1.5. Although the FID score slightly increases, the overall difference is not significant according to the visualizations.

\begin{figure*}[t]
\centering
\includegraphics[width=1\textwidth]{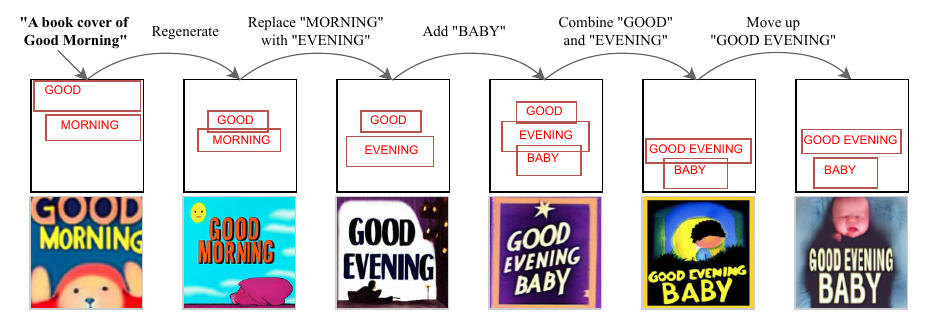}
\caption{
Visualizations of operating layout through multi-round chat and the corresponding images.
}
\vspace{-0.4cm}
\label{fig:link}
\end{figure*}

\subsection{Discussions}

\paragraph{Operating layout through multi-round chat.} Since the language model used to generate the layout is fine-tuned based on a chat model, we are curious whether we can manipulate the layout through multi-round chat. We demonstrate the results in Figure  \ref{fig:link}. Experimental results showcase that through interactive conversation, $\mathbf{M_{1}}$ can not merely regenerate the layout, but also possess the ability to add or modify keywords, as well as manipulate the location of the box. This further enhances the flexibility and versatility of the proposed TextDiffuser-2.

\paragraph{Generation based on overlapping layouts.} Occasionally, we notice that there exist overlapping boxes during the layout prediction stage. We present TextDiffuser-2, as well as the results generated by GlyphControl and TextDiffuser using overlapping layouts in \cjy{Appendix I}. Experimental results indicate that TextDiffuser-2 demonstrates greater robustness towards overlapping boxes. Conversely, the results generated by the other two methods will produce scrambled text, thereby impacting the overall quality of the image.

\section{Conclusion}
In this paper, we introduce TextDiffuser-2, aiming to unleash the power of language models for the text rendering task. Specifically, we attempt to tame two language models, one for layout planning and the other for layout encoding. Experimental results validate that TextDiffuser-2 is capable of generating more diverse images while maintaining the accuracy of the generated text. For the \textit{limitation}, TextDiffuser-2 faces challenges when rendering complex languages, as it expands the renderable character table by adding new tokens. For instance, when rendering Chinese text, TextDiffuser-2 faces difficulties due to the extensive character set, potentially leading to few-shot or even zero-shot scenarios. For the \textit{broader impact}, TextDiffuser-2 can be used to enhance creativity in fields like graphic design, advertising, and art. Besides, it can be used to generate informative images for teaching and learning. For example, it could create diagrams with explanatory text
for educational materials. However, if used maliciously, TextDiffuser-2 could be employed to create images containing false text information. For \textit{future work}, we seek to explore the rendering of characters in multiple languages and enhance the resolution of generated text images.

\appendix

\noindent{\Large{\textbf{Appendix}}}



\section{Choice of the maximum length $L$}
\begin{figure}[h]
\centering
\includegraphics[width=0.5\textwidth]{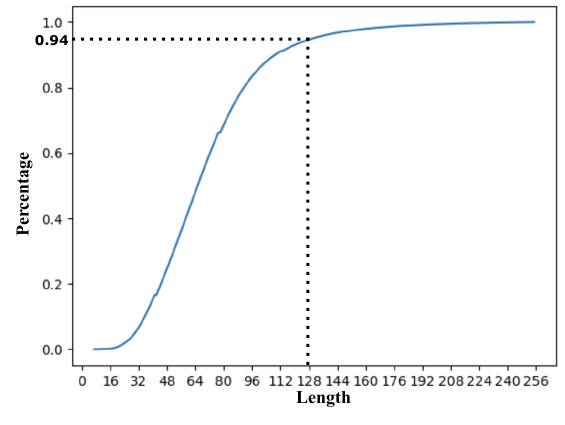}
\caption{The cumulative distribution function to analyze the length of the composed sequences. When setting the maximum length $L$ at 128, the vast majority of samples (94\%) are covered.}
\label{fig:cdf}
\end{figure}

During the training process, the composed sequence (\textit{i.e.}, the prompt combined with text content and position) has a maximum length limit. As shown in Figure \ref{fig:cdf}, by analyzing the MARIO-10M dataset \cite{chen2023textdiffuser}, we notice that the composed sequence for 94.0\% of the samples is less than 128 in length, and all samples are below this threshold during the evaluation. Obviously, we can increase the maximum length to a larger value, such as 256, accommodating 99.2\% of training samples. However, enlarging the length limit would also result in increased computational costs, such as raising the single sample inference time from 6 to 7 seconds. Therefore, the choice of length limit should be made based on practical considerations, balancing between the model's capability and efficiency.

\section{More visualizations of layout prediction}

\begin{figure}[t]
\centering
\includegraphics[width=0.48\textwidth]{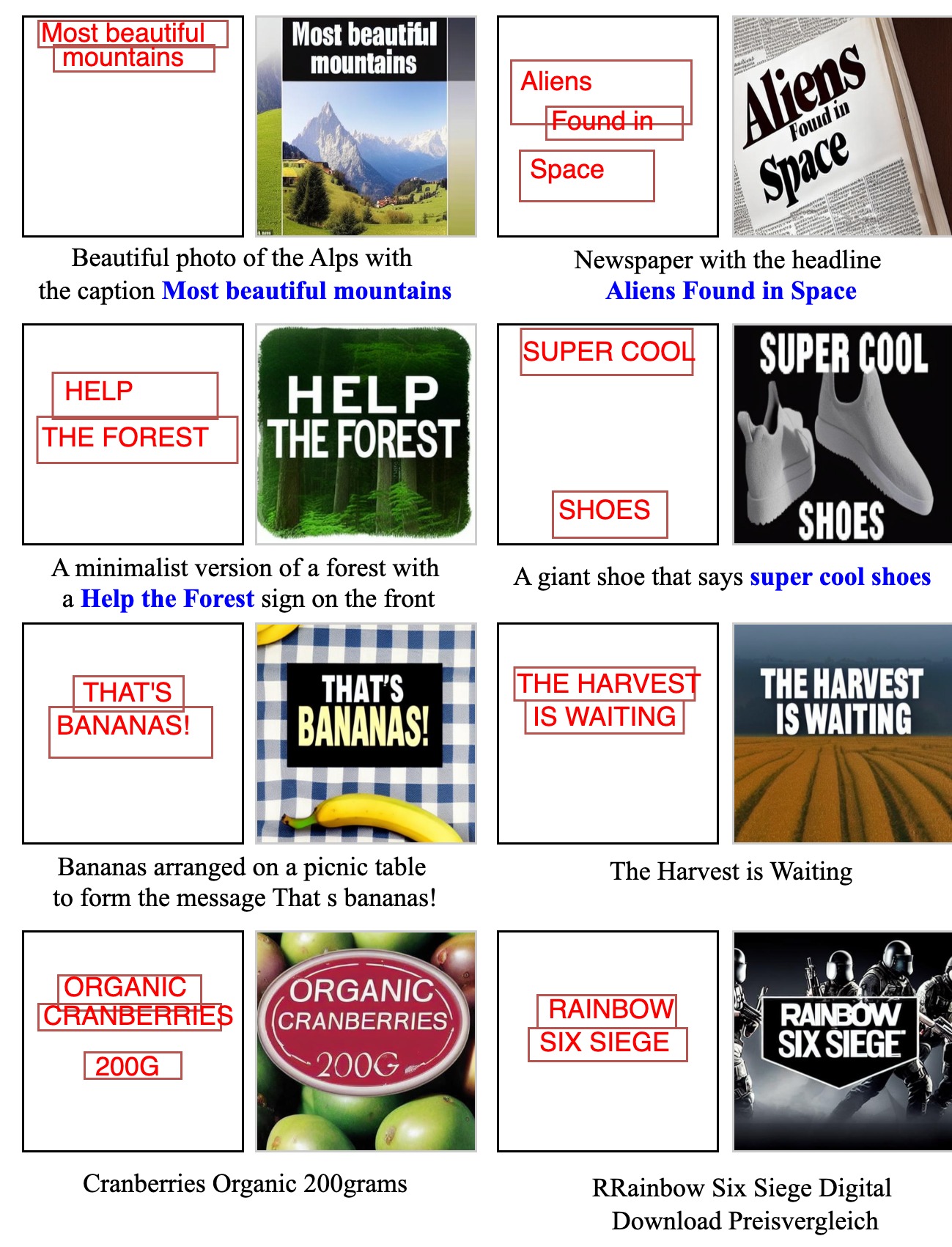}
\caption{More visualizations of the layout predictions. The specified keywords are marked in blue color in the first two rows.}
\label{fig:morelayout}
\end{figure}

As depicted in Figure \ref{fig:morelayout}, we showcase more layout prediction results. We specify the keywords to be rendered in the first two rows. The language model has the capacity to organize the specified keywords, placing related keywords in the same line and generating aesthetically pleasing layouts. Notably, the final row of predictions includes words not present within the prompt. For instance, the model substitutes ``200g'' for ``200gram''. It is a logical substitution given that both terms convey the same meaning. Additionally, the model replaces the misspelled term ``RRAINBOW'' in the prompt with the correct term ``RAINBOW''. This further showcases the flexibility of the layout planner $\mathbf{M_{1}}$.

\begin{figure*}[t]
\centering
\includegraphics[width=1\textwidth]{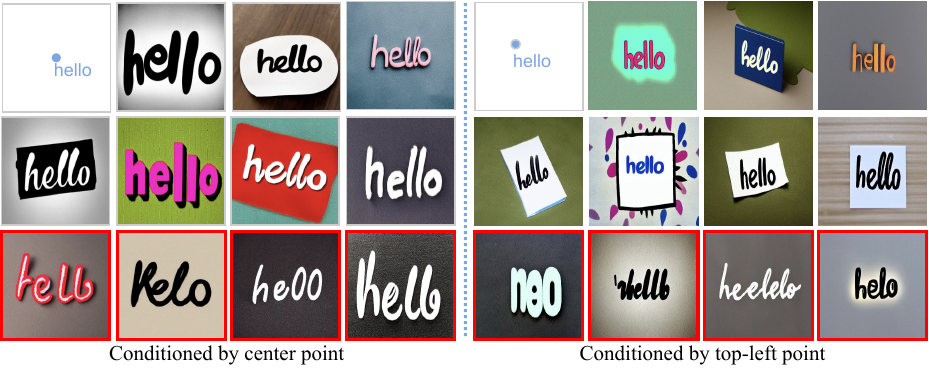}
\caption{Visualizations of generation guided by single-point conditions, including the center point and the top-left point. The prompt is ``A text image of hello''. The samples highlighted by red boxes in the last row denote the rendered text is incorrect.}
\label{fig:singlepoint}
\end{figure*}

\begin{figure*}[t]
\centering
\includegraphics[width=1\textwidth]{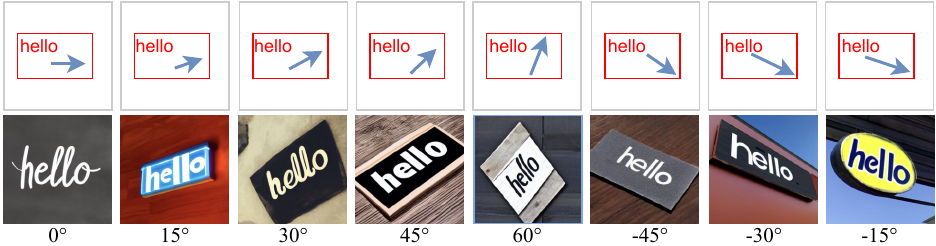}
\caption{Visualizations of generation with different angle guidance. The prompt is ``A text image of hello''.}
\label{fig:angle}
\end{figure*}

\begin{figure*}[t]
\centering
\includegraphics[width=1\textwidth]{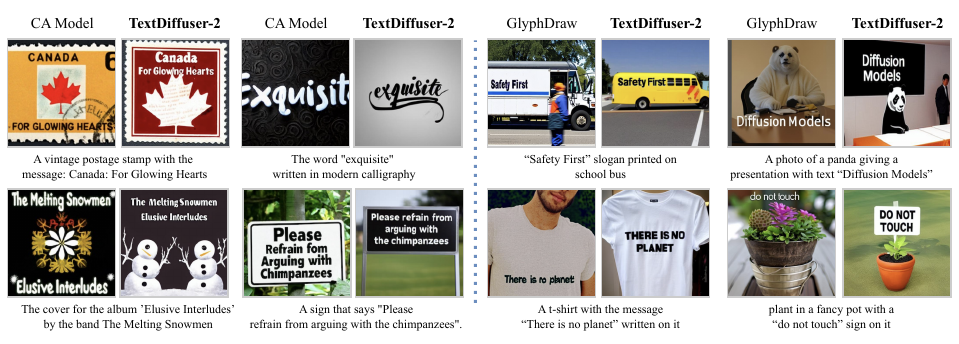}
\caption{Comparisons with Character-Aware Model (CA Model) \cite{liu2022character} and GlyphDraw \cite{ma2023glyphdraw} using samples in their papers.}
\label{fig:other}
\end{figure*}

\section{Generation guided by single-point condition}

We retrain TextDiffuser-2 and implement a single-point supervision strategy during the training process, such as using the center and top-left points. As illustrated in Figure \ref{fig:singlepoint}, despite the diversity in text size and angle generated by the single-point conditions, we observe a significant portion of the text to be inaccurate. Given the observation of a decline in accuracy over 20\% (as shown in Table 2 in the main paper), we ultimately employ the top-left and bottom-right points as the condition.

\section{Generation with additional angle condition}
As shown in Figure \ref{fig:angle}, we demonstrate samples generated with different angle conditions. Specifically, we retrain TextDiffuser-2 and add 181 angle tokens, ranging from -90$\degree$ to 90$\degree$. When constructing the language-format layout, the angle token is placed after the four coordinate tokens. The results show that the generated results align well with the angle instructions. 

\section{Details of compared methods and evaluation}
We introduce all the baselines and their experimental settings as follows.

\noindent \textbf{SD-XL} \cite{podell2023sdxl} is an improved version of the latent diffusion model \cite{rombach2022high} with stronger backbone and powerful text embedding. SD-XL comprises 5.8B parameters and the resolution of the output images is 1024$\times$1024.

\noindent \textbf{PixArt-$\alpha$} \cite{chen2023pixart} is a powerful Transformer-based text-to-image diffusion model and is training-efficient. It consists of 0.6B parameters. The output resolution is of size 1024$\times$1024.

\noindent \textbf{Ideogram} \cite{ideogram} is an online website that can produce attractive logos, posters, and other natural images based on prompts. We use the typography mode and manually quote keywords to be rendered. The resolution is 1024$\times$1024.

\noindent \textbf{DALLE-3} \cite{dalle3} exhibits robust text-to-image capabilities, producing images that precisely conform to the given prompt. It generates high-resolution outputs with 1024$\times$1024 resolution. We leverage the official API for the generation process.

\noindent \textbf{GlyphControl} \cite{yang2023glyphcontrol} utilizes the framework of ControlNet \cite{zhang2023adding} and the pre-trained model of SD 2.1 \cite{rombach2022high}, producing the output image of size 768$\times$768. It takes glyph images with multiple text lines as the condition. GlyphControl has 1.3B parameters. Specifically, since it can not generate images from prompts, we use the layouts produced by TextDiffuser-2 to make the glyph image.

\noindent \textbf{TextDiffuser} \cite{chen2023textdiffuser} is a two-stage framework that can convert user prompts into images. It relies on users to specify keywords for rendering. TextDiffuser is pre-trained based on SD 1.5 \cite{rombach2022high}, and the resolution of the generated images is 512$\times$512. It consists of 884M parameters in total.

For evaluation, we utilize the metrics employed in TextDiffuser \cite{chen2023textdiffuser} and also use \href{https://learn.microsoft.com/en-us/azure/cognitive-services/computer-vision/how-to/call-read-api}{\textcolor{black}{Microsoft Read API}} to evaluate the OCR performance.

\section{Comparisons with samples in other papers} 
Since the source code, pre-trained weight, or demo is not available for Character-Aware Model \cite{liu2022character} and GlyphDraw \cite{ma2023glyphdraw}, we conduct comparisons with samples in their corresponding papers. As demonstrated in Figure \ref{fig:other}, we visualize four samples for each compared method. Notably, TextDiffuser-2 shows better rendering accuracy compared with the Character-Aware Model, which contains several typos, including the missing ``r'' in ``from'' and the incorrect spelling of ``Chimpanzees''. Besides, the Character-Aware Model enhances visual text rendering by utilizing language models with a larger parameter size (\textit{e.g.}, T5-XXL \cite{raffel2020exploring} with 11B parameters). We have demonstrated that even with a smaller-scaled CLIP text encoder with 63M parameters, superior text rendering performance can be achieved by virtue of explicit positional and content supervision. Besides, TextDiffuser-2 outperforms GlyphDraw as TextDiffuser-2 can render images with multiple text lines.

\section{More details about user studies}

\begin{figure*}[t]
\centering
\includegraphics[width=0.78\textwidth]{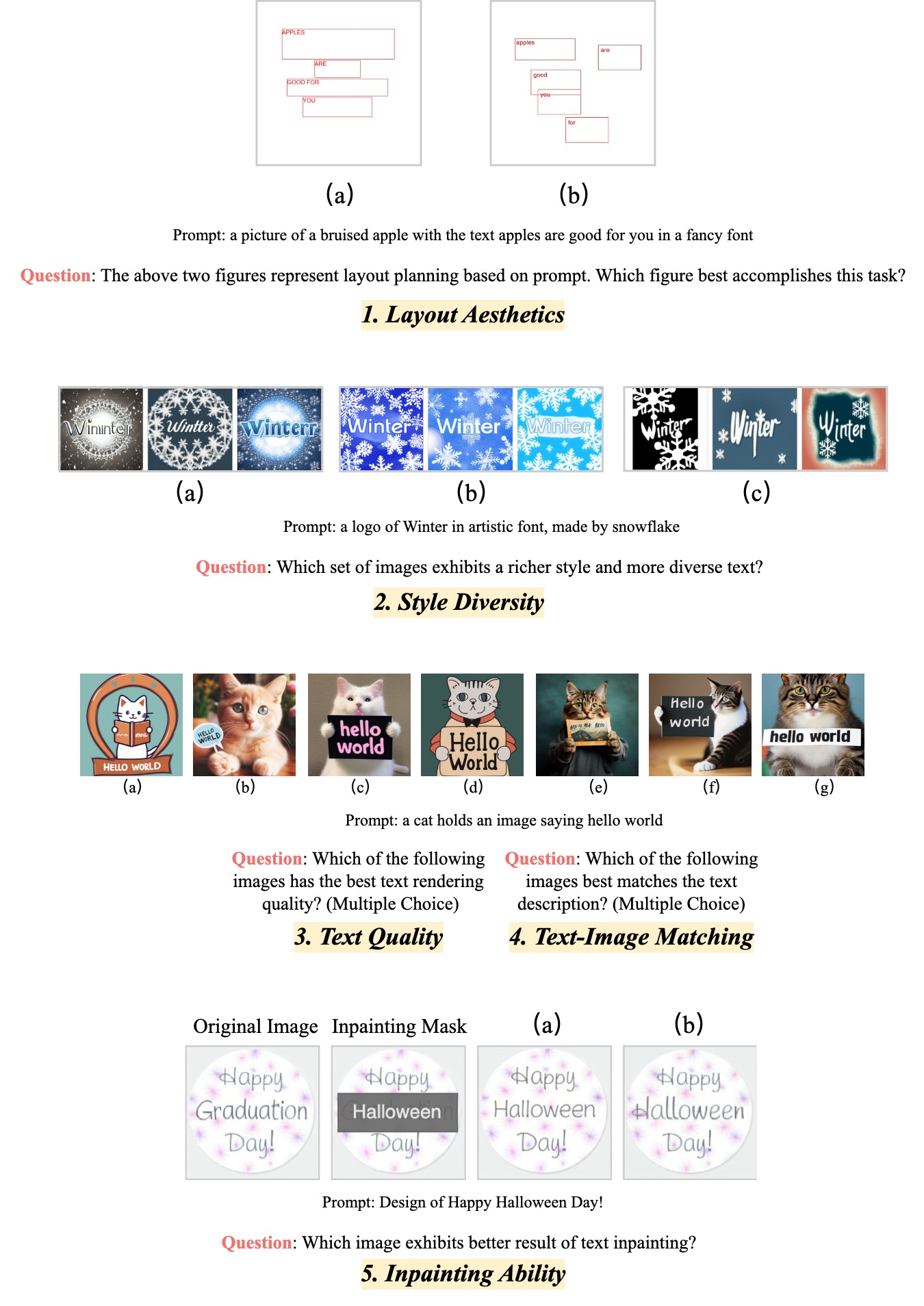}
\caption{The user studies comprise five questions. Specifically, users should make a single selection for questions pertaining to layout aesthetics, style diversity, and inpainting ability. For questions regarding text quality and text-image matching, users have the freedom to select multiple options.}
\label{fig:user}
\end{figure*}

\begin{figure*}[t]
\centering
\includegraphics[width=0.71\textwidth]{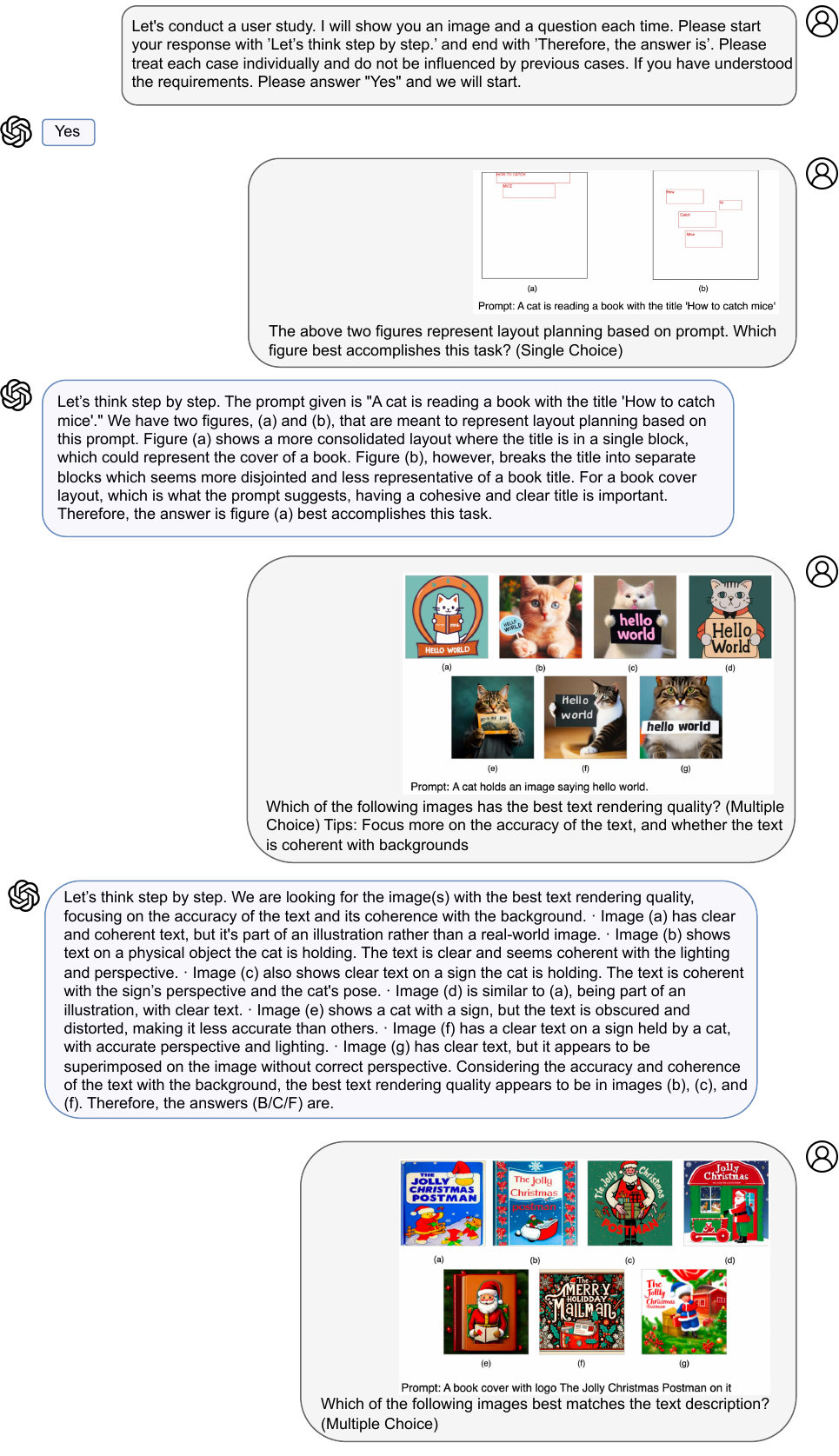}
\caption{The illustration of using GPT-4V \cite{gpt4} to conduct user studies (Pages: 1/2).}
\label{fig:gpt1}
\end{figure*}

\begin{figure*}[t]
\centering
\includegraphics[width=0.71\textwidth]{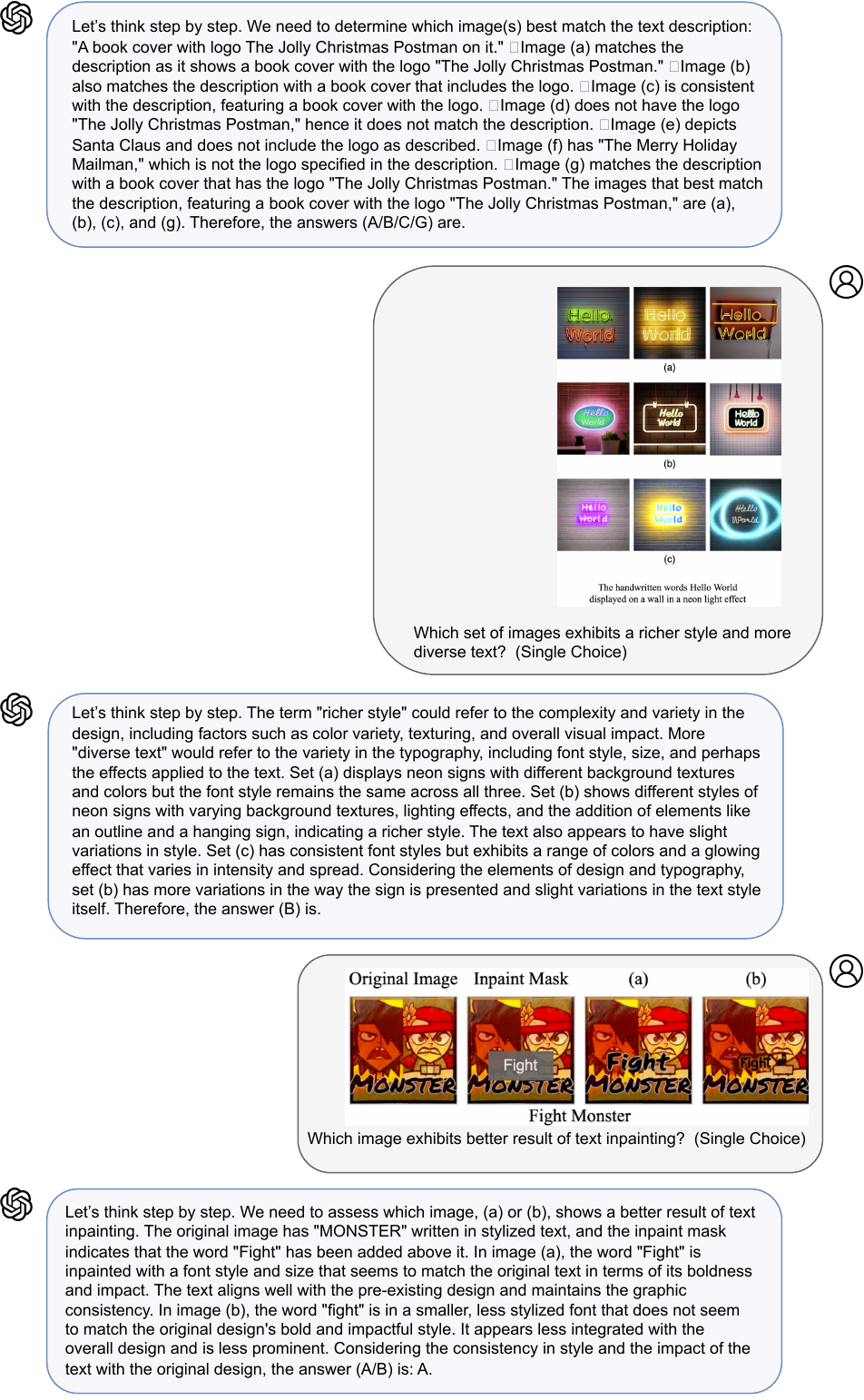}
\caption{The illustration of using GPT-4V \cite{gpt4} to conduct user studies (Pages: 2/2).}
\label{fig:gpt2}
\end{figure*}

We conduct comprehensive user studies on five aspects, including layout aesthetics, style diversity, text quality, text-image matching, and inpainting ability. The details of the questions are displayed in Figure \ref{fig:user}. In addition to human involvement, we incorporate GPT-4V \cite{gpt4} in our user studies. Specifically, we design prompts to encourage GPT-4V to proceed step-by-step, deriving the final answer through logical analysis. The dialogue record is shown in Figure \ref{fig:gpt1} and Figure \ref{fig:gpt2}. It suggests that GPT-4V exhibits impressive literacy skills, and its logical chain is reasonable.

\section{More details about text inpainting}
Similar to TextDiffuser \cite{chen2023textdiffuser}, by appending another five-dimension feature, including the one-dimension mask and four-dimension non-inpainted area features, to the input of U-Net, TextDiffuser-2 can be trained for the text inpainting task. Specifically, 14,400 parameters will be added, which accounts for a small proportion of the whole architecture containing 922M parameters. We set the classifier-free guidance to 7.5 and used 50 sampling steps, which cost 6 seconds for generation using one A100 GPU card.

\section{Generation based on overlapping layouts}

\begin{figure}[t]
\centering
\includegraphics[width=0.48\textwidth]{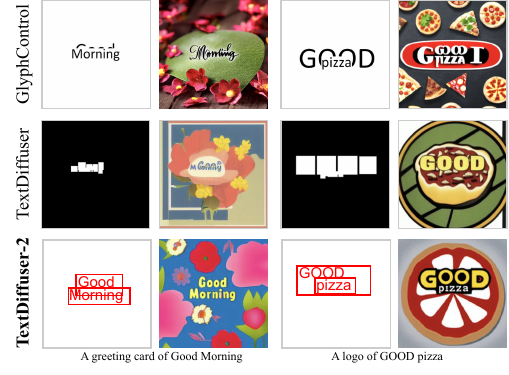}
\caption{Comparative visualizations of generation results using overlapping layouts. TextDiffuser-2 demonstrates enhanced robustness compared with other methods.}
\label{fig:overlap}
\end{figure}

The results of the layout predictor will inevitably contain overlapping boxes. In this section, we analyze the robustness of three methods, including GlyphControl \cite{yang2023glyphcontrol}, TextDiffuser \cite{chen2023textdiffuser}, and the proposed TextDiffuser-2. In terms of explicit guidance, GlyphControl employs glyph images, TextDiffuser uses character-level segmentation masks, and TextDiffuser-2 harnesses bounding boxes with corresponding text. The visualization results are demonstrated in Figure \ref{fig:overlap}. The results reveal that the proposed TextDiffuser-2 is more robust when using overlapping layouts for a generation. By contrast, GlyphControl and TextDiffuser will generate incorrect text, resulting from the occluded glyph images and segmentation masks.

\begin{figure}[t]
\centering
\includegraphics[width=0.48\textwidth]{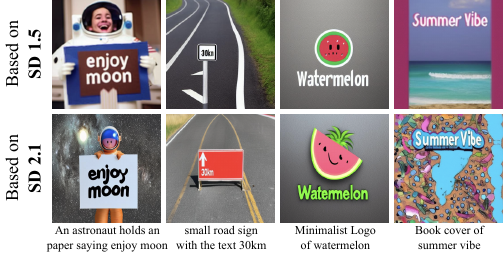}
\caption{Comparative visualizations of TextDiffuser-2 with different versions of the Stable Diffusion model. Utilization of SD 2.1 exhibits improved detail rendering and a more accurate depiction of small-scale characters compared to SD 1.5.}
\label{fig:21}
\end{figure}

\begin{figure}[t]
\centering
\includegraphics[width=0.48\textwidth]{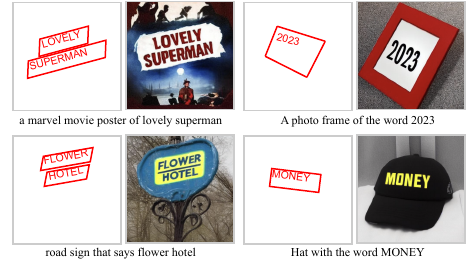}
\caption{Demonstration of TextDiffuser-2's generation guided by quadrilateral bounding boxes, showcasing the model's ability to align text accurately within the specified geometrical constraints.}
\label{fig:quad}
\end{figure}

\section{Conformity of the positional guidance}
To verify whether the text generated in the images adheres to positional guidance, we employ a widely-used text detection tool \cite{sun2019icdar} for assessment. The precision, recall, and F-measure are 0.9524, 0.9635, and 0.9579, respectively. This indicates that in the vast majority of cases, the generated text is able to comply with the provided positional constraints.

\section{TextDiffuser-2 based on SD 2.1}
TextDiffuser-2 can be trained based on different pre-trained models. In Figure \ref{fig:21}, we visualize the comparisons of results based on SD 1.5 and SD 2.1. We notice that the model based on SD 2.1 provides more details thanks to the stronger power of the pre-trained model. Meanwhile, it can correctly render characters with small sizes because the resolution of the latent space is higher.

\section{Generation guided by quadrilateral boxes}
In addition to using horizontal boxes to provide positional information, we remain curious whether TextDiffuser-2 can be guided by quadrilateral boxes, which could more accurately describe slanted text. To investigate this, we make two modifications. First, we train a layout planner $\mathbf{M1}$ to output each line in the format of ``\textit{textline} $x0,y0,x1,y1,x2,y2,x3,y3$''. Secondly, we adapt the layout encoder $\mathbf{M2}$ to encode this sequence. We set the maximum length limit $L$ to 256 to accommodate longer input sequences. Visualizations are shown in Figure \ref{fig:quad}. We notice that the generated results align well with the guidance of quadrilateral boxes. For future work, we plan to use more control points to represent the boxes, allowing for rendering more artistic text.

\clearpage
\clearpage

{
    \small
    \bibliographystyle{ieeenat_fullname}
    \bibliography{main}
}

\end{document}